# Low Speed Automation, a French Initiative

Sébastien GLASER, Maurice COUR, Lydie NOUVELIERE, Alain LAMBERT, Fawzi NASHASHIBI, Jean-Christophe POPIEUL, Benjamin MOURLLION

*(respectively)*
IFSTTAR, 14 route de la minière, 78000 Versailles - FRANCE
CONTINENTAL Automotive SAS, 1 avenue Paul Ourliac, 31036 Toulouse - FRANCE
IBISC - UEVE, 40, rue du Pelvoux,91020 Evry  - FRANCE
IEF, Bat 220, centre scientifique d'Orsay, 91405 Orsay -FRANCE
INRIA,Domaine de Voluceau BP 105, 78153 Le Chesnay  - FRANCE
LAMIH, UVHC, Le mont Houy, 59313, Valenciennes -FRANCE
MIPS, 2 rue des frères Lumière,68093 Mulhouse -FRANCE

**Abstract**

Nowadays, vehicle safety is constantly increasing thanks to the improvement of vehicle passive and active safety. However, on a daily usage of the car, traffic jams remains a problem. With limited space for road infrastructure, automation of the driving task on specific situation seems to be a possible solution. The French project ABV, which stands for low speed automation, tries to demonstrate the feasibility of the concept and to prove the benefits. In this article, we describe the scientific background of the project and expected outputs.

## 1. Introduction

Vehicle safety and Traffic jams are recurring problems in all the major cities. The improvement of vehicle passive and active safety, of the infrastructure design and of driver behavior, makes it possible to reach very low number of car fatalities. Several driving assistances enhance the controllability of the vehicle as anti-lock brake system (ABS), electronic stability program (ESP) … can support the driver as Adaptive Cruise Control (ACC) and even can decide to brake in front of an obstacle that is not avoidable in the case of a collision mitigation brake system. However, even with improved public transportation systems, the traffic jams problem remains and increases because of an increased daily traveling, of a need of mobility…

Previous works show that, with a limited space to increase the road possibility, the automation of the vehicle could be a solution. Several experiences were driven in the US, in Japan and in European countries during the 90's and in the beginning of 2000. They prove that the full automation is possible but at an expensive price. During the past five years, several challenges, as DARPA challenges, confirm the possibility of the automation. Automates are able to drive on road that ranges from highway to urban area. Moreover, recent developments in the automotive sensors, actuators and micro controllers, make the automation possible at a reasonable price. On recent cars, where driving assistances are available, plenty of sensors can detect objects in the surrounding, and controllable actuators are available at serial production level. But, some actuators, especially related with the steering column, are not usual as they are strongly limited by current legal issue.

However, the developed applications miss the link with the driver. They are mostly automates that take or not the control of the car. The relation with the driver is limited. European projects as SPARC (SPARC 2002) try to develop safe architecture to allow a fine tuning of the control between the driver and the driving assistance. More recently, the HAVEit project (HAVEit 2008) has developed a copilot architecture that is able to monitor the driving task and to interact with him to initiate the control by an automate. The French project PARTAGE (PARTAGE 2009) studies several scenarios to share the control between a driving assistance and the driver, from the warning to the full automation to provide a full scale of interaction.

In ABV, Low Speed Automation, a French initiative, we focus on the problem of traffic jam in urban area and we try to provide a solution for ring road, or interurban freeway under mixed traffic condition. We define a vehicle that is able to handle partly or totally the driving task on a secured road but taking into account interactions with others vehicles:
- Automation: The system will provide up to a full control of the driving task, managing also the transition between the driver and the automation.
- Low speed: the full automation of the vehicle will only be available up to 50km/h. For a speed higher than 50km/h, the driver is still assisted but only for the longitudinal mode, providing functions like an ACC
- Secured Road: a secured road is not a road available only for 'ABV' vehicles; the road allows a shared driving space between automated vehicle and normal vehicle. A secured road provides a high quality of services, as a good road marking, knowledge of the infrastructure geometry…

To allow the shared control, the project uses copilot architecture in interaction with the driver and also monitors him to interact during transition phases (between the driver and the copilot).

ABV project is structured along three main tasks:
- Scientific development: it gathers the development of the perception, the path planning, the control and the cooperation between the human and the machine,
- Integration in the vehicle: it aims at developing the specification and the architecture of the vehicle, at the evaluation of the modules before the integration and finally at the integration within two demonstrators,
- Societal aspect: in order to evaluate the impact of the project at a larger scale than the vehicle and also to determine the legal evolution.

ABV gathers 10 French partners with equipment supplier and large industry, sme, research units and university. The project starts two years ago and we present here the first results. We present here the first results.

After defining the basic scenario, we focus on the perception, data fusion and control of the vehicle in interaction with the driver. Two prototypes are developed and they allow this interaction based on by-wire architecture. Moreover, the project also studies the impact on the traffic flow, on global consumption and the constraints in term of legal aspect.

**2. Scenarios for ABV**

In the frame of ABV, we make the following distinction for the available functions:
- Speed: a threshold on the speed is set at 50km/h. Below 50km/h the ABV functions could be fully available. For a speed higher than 50km/h, the driver is assisted only for the longitudinal mode.

- Road: if the road is a secured road, then the ABV functions could be fully available. On normal road, depending on the speed, the drivers could be assisted on longitudinal mode.

The threshold on speed is set at 50km/h mainly for sensing purpose: from previous experience (DO30 and LOVE projects), we can ensure the obstacle detection using a camera based system up to 40m in front of the vehicle. At 50km/h, this distance allows a reaction time greater than 2s. The system can then safely warn the driver and/or decides on appropriate reaction within this time. Moreover, without any reaction from the driver, we can fully stop the vehicle.

*2.1. Scenarios*

From the previous limitation, we have defined six different behaviors for the ABV systems which are presented on the Figure 1.

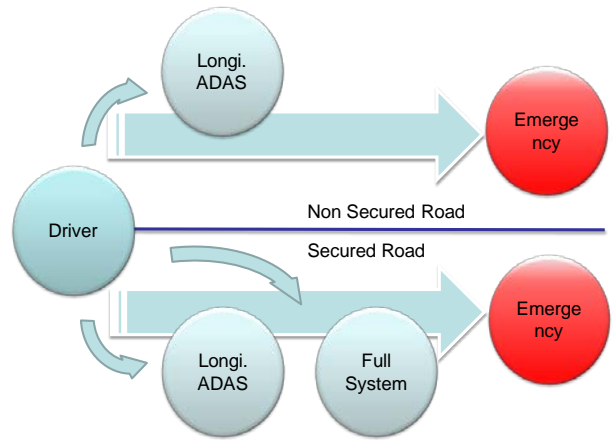

Figure 1 Behavior of the ABV system

These basic behaviors can be described as:
- "Driver" in this mode, the driver fully controls the vehicle
- "Longi ADAS" mode is available both on secured road and on non secured road. In this mode, the system supports the driver for the longitudinal decision.
- "Full system" mode is available only on Secured Road. The driver can fully delegate the control of the vehicle to the system. This mode is only available if the speed is below 50km/h.
- "Emergency" mode aims at stopping the vehicle in a safe state if a failure is detected. On non secured road, the aim is to stop the vehicle on the lane, on secured road, the system tries to stop the vehicle on the emergency lane.

Each mode can be disengaged by the driver and the driver, on Secured Road, can be proposed by the driving assistance to engage the system.

The "Emergency" mode can be triggered autonomously. For instance, if the system is on "Full System" mode, and that the Secured Road ends, the driver will be asked to take the control back. If the monitoring system does not recognize any reaction from the driver, then the system switches to the "Emergency" mode.

*2.2. From scenarios to architecture and functions*

From the previous scenarios, the functions that must be achieved by the vehicle are the following ones:
- Perception and Data Fusion: Positioning of the vehicle, lane detection,
- Path Planning
- Driver Monitoring
- Human Machine Interface
- Vehicle control

The architecture is summarized in the Figure 2.

*2.3. Secured Road Concept*

As in the SMART final report (SMART 2011), we clearly think that the vehicle alone cannot achieve a level of reliability that is sufficient to allow autonomous system on open road. These systems will need cooperation with, at least, the infrastructure and the other vehicles. This is the concept of a "Secured Road".

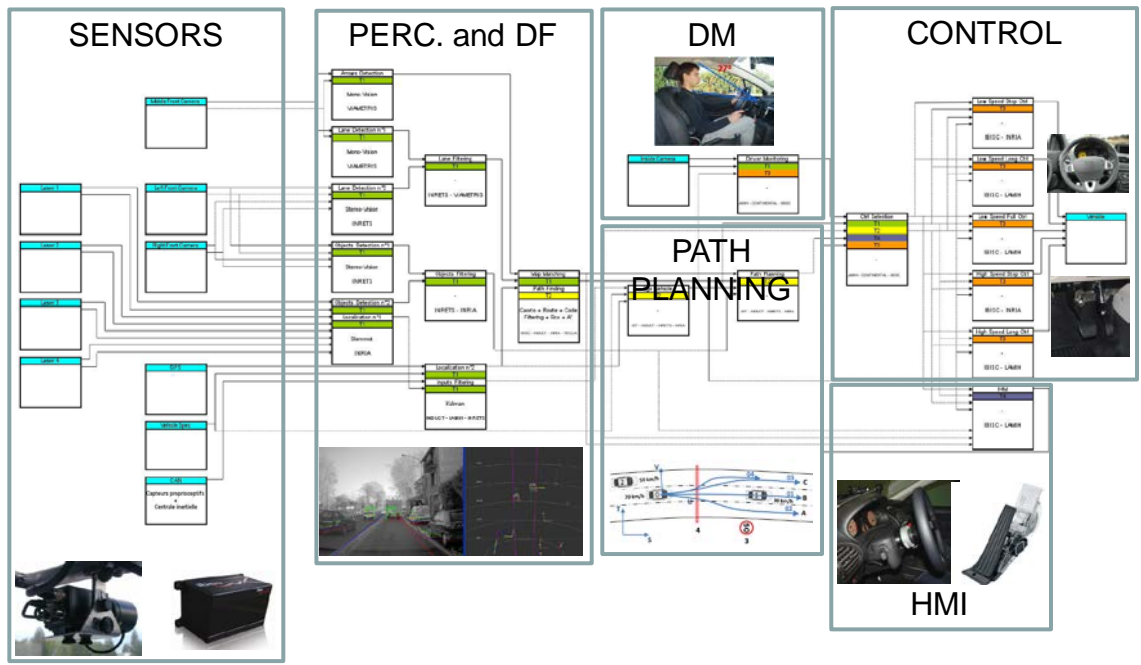

Figure 2 Description of the architecture

A "Secured Road" is a road that accepts mixed traffic: both ABV vehicles and conventional vehicles. This infrastructure primarily aims at easing the understanding of the environment. According to the described architecture, the sensing functions have to detect the road and the obstacle. The infrastructure can help in these two cases. For the road, the infrastructure may ensure that the road marks and the police sign could be easily detected (i.e. not deteriorated) or that an accurate local map is available for the vehicle with the data from the road and static objects. For the obstacle, even if solutions to detect obstacles from the infrastructure are available (i.e. laserscanners and V2I communication), the price may limit the development. However, it is interesting to have sensing capacities where a lot of interaction may occur, as in intersection.

Moreover a "Secured Road" is also able to supervise its use. Gathering information from vehicles and also static sensors, the infrastructure is able to have a global overview of its current status and distribute specific recommendation, as speed limit or dedicated trip to ABV vehicles in order to optimize the traffic.

**3. Details of the architecture**

*3.1. Sensing the environment*

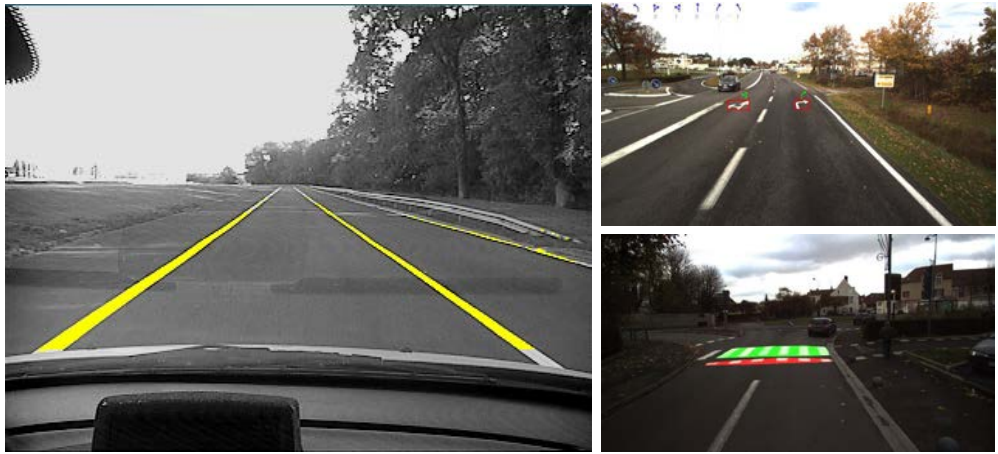

Figure 3 Road detection

This is the first functions block. It aims at providing a map of the surrounding of the vehicle. Both lanes and obstacle must be known. Multi-Lane detection is done using multi cameras architecture (Pollard 2011). Moreover, the road specific elements as arrows or pedestrian ways are detected in order to locally modify the behavior of the others functions as path planning and vehicle control. Some results are presented in the Figure 3. The object detection is actually done using fusion between laserscanner and camera (Labayrade 2007 and Labayrade 2010).

*3.2. Generating a set of trajectories*

The core concept of the trajectory generation is that the vehicle must obey to the Vienna convention. Moreover, we make the assumption that the other drivers do not any actions that can endanger the surrounding vehicle (Vanholme 2011). Then, the generation of the trajectories is done using several competitive methods:
- Polynomial trajectory definition under constraint (S. Glaser 2010). This method is fast and reliable. It is already implemented under simulator and also on car.. This method was partly developed during the HAVEit project. The Figure 4 shows what are the considered trajectories for the other vehicles while defining the ego vehicle trajectory)
- Geometric with uncertainty and Bayesian approach with uncertainty are currently under development.

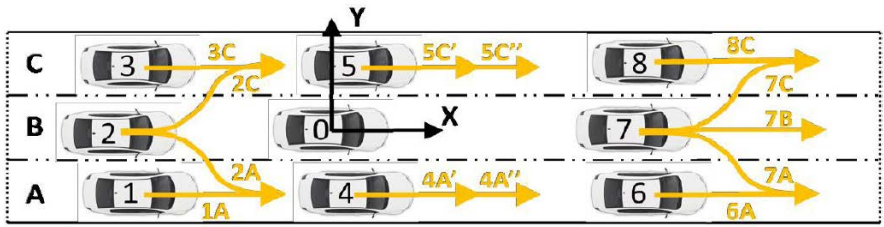

Figure 4 Considered trajectories for the ego vehicle trajectory generation

*3.3. Controlling the vehicle*

Once the environment sensed, and the trajectory generated, the control module aims at providing the motion vector (in term of steering angle and speed, see Figure 5) and also to optimize the consumption of the vehicle while following the trajectory (Nouveliere 2011).

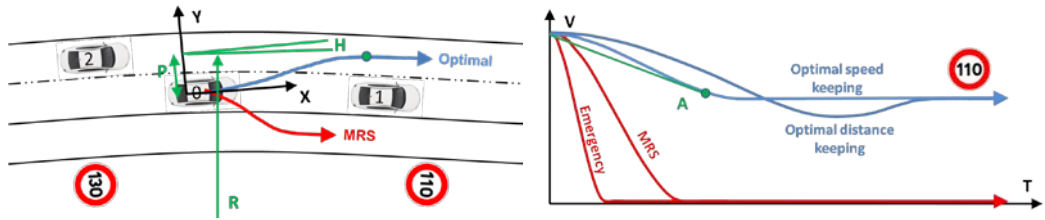

Figure 5 Generation of the commands. Two trajectory and command associated are continuously evaluated : Minimum Risk State (MRS) and emergency

The minimum risk state (emergency control while on "Secured Road") and the emergency trajectory are continuously evaluated.

*3.4. Cooperation between man and machine*

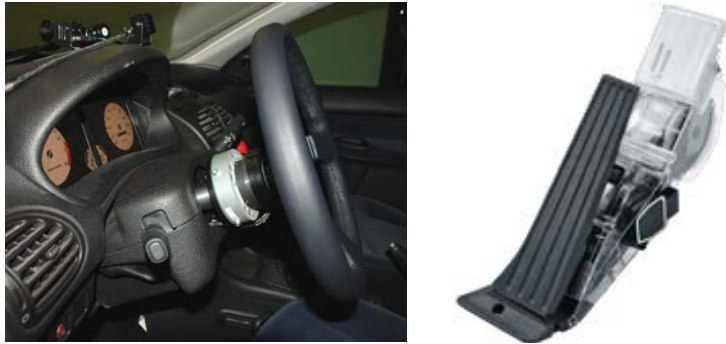

Figure 6 By wire Steering wheel and active pedal

In order to ensure a good understanding from the driver, and a possible sharing of the control between the driver and the automate, we have specifically addressed this topic in the project. The steering wheel and column are transformed to be by-wire, and the driver feedback is defined to allow a safe feeling (Sentouh 2010, Soualmi 2010, Sentouh 2011). The main idea is that the driver feels a self aligning torque that brings the vehicle softly back to the trajectory when the control is shared. However, he can override the command and the torque is maintained at small values. The steering column is not the only way to share the control with the driver; the acceleration pedal is also designed to be able to generate a feedback to the driver. This modality is not yet explored. The Figure 6 presents these interfaces. Moreover, the driver is monitored to ensure that he is able to take the control back and that he is aware of the situation. The work is actually done under a simulator (Figure 7).

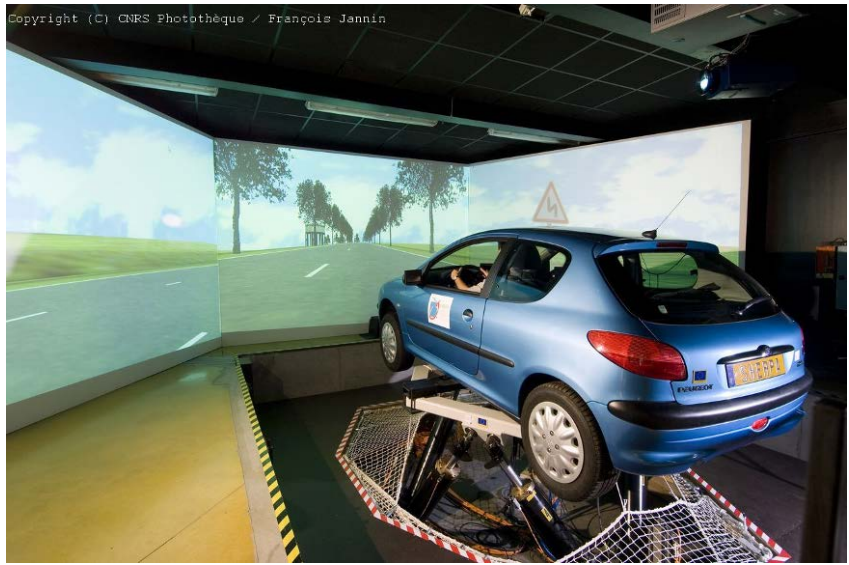

Figure 7 SHERPA simulator at LAMIH university

## 4. Societal aspect

Beyond the technical aspects, the project also aims at evaluation the impact on societal aspect. On this part, the project aims at:
- Studying what are the legal limitations that forbid full autonomous systems on open road and design possible evolution, both on the system and on the legal aspect.
- Evaluating the impact of the system on the traffic with a scaled introduction of the system in the traffic.

Especially for this last part, the objectives are to evaluate the impact not only on the mobility, but also on the safety and on the consumption. For this last criterion, the model will be developed in collaboration with the control part of the project.

## 5. Conclusion and perspectives

In this paper, we have presented the structure and the first results of the ABV project. This project aims at proposing a solution that enhances both safety and mobility by developing two demonstrators of cars that are able to share the control between the driver and the automation up to the full automation on a "Secured Road" when traffic jams occur. The project aims at developing integrated demonstrators that promote the autonomous driving.

Moreover, we not only study the autonomous vehicle, but all the possible transitions that ranger from warning to highly assisted.

The main objective is to bring a solution on highway and urban road that are prone to traffic jams, by developing the concept of "Secrued Road", a road that provides a high quality of services and that allow and help the automation in the detection and understanding phases.

The project will end at the beginning of 2013, we are currently integrating and evaluating the functions.

## Acknowledgements

The authors would like to thanks the French National Research Agency (ANR) to fund the ABV project and allowing this research. ABV project leader is IFSTTAR and partnership includes : CONTINENTAL Automotive France, IBISC - UEVE, GM Conseil, IEF, INDUCT, INRIA, LAMIH, VIAMETRIS, MIPS and VERI.